\makeatletter
\newcommand\notsotiny{\@setfontsize\notsotiny\@vipt\@viipt}
\makeatother
\documentclass{bvm} 
\begin{document}

\newcommand{\bvmyear}{2026}

\selectlanguage{english} 

\title{Few-Shot Fingerprinting}

\subtitle{Subject Re-Identification in 3D-MRI and 2D-X-Ray}

\titlerunning{F2F: Subject ReID}

\author{
    Gonçalo Gaspar Alves\inst{2},
    Shekoufeh Gorgi Zadeh\inst{1},
    Andreas Husch\inst{2},
    Ben Bausch\inst{1}
}

\authorrunning{Gaspar et al.}

\institute{
\inst{1} University of Luxembourg
\inst{2} Former Association with University of Luxembourg
}

\email{ben.bausch@uni.lu}

\maketitle

\begin{abstract}
Combining open-source datasets can introduce data leakage if the same subject appears in multiple sets, leading to inflated model performance. To address this, we explore subject fingerprinting, mapping all images of a subject to a distinct region in latent space, to enable subject re-identification via similarity matching. Using a ResNet-50 trained with triplet margin loss, we evaluate few-shot fingerprinting on 3D MRI and 2D X-ray data in both standard (20-way 1-shot) and challenging (1000-way 1-shot) scenarios. The model achieves high Mean-Recall-@-K scores: 99.10\% (20-way 1-shot) and 90.06\% (500-way 5-shot) on ChestXray-14; 99.20\% (20-way 1-shot) and 98.86\% (100-way 3-shot) on BraTS-2021.
\end{abstract}

\section{Introduction}

In our context, subject re-identification does not aim to reveal the subject's identity from anonymized data, as described in \cite{packhauser2022deep}, but identifying all images that belong to the same subject. Therefore we find the term subject fingerprinting more suitable. This technique leverages anatomical structures and pathophysiological conditions as unique markers to associate an image with a specific individual. This formulation is very useful in dataset aggregation, where we need to identify subjects present in multiple datasets in order to avoid data leakage.\\
Some methods for subject reidentification in MRI, such as \cite{Truong2024Benchmarking}, use majority voting comparing multiple $2$D slices from a pair of $3$D MRI volumes to detect duplicate or near-duplicate volumes. Similar to our work, \cite{macpherson2023patient} and \cite{Packh_user_2022} compare latent space embeddings of multiple images to determine if they belong to the same subject. To generate these image embeddings, researchers commonly apply deep neural networks to encode the images to fixed size embedding vectors. Training such models can be done in an unsupervised manner, as the only form of supervision needed is knowing which images belong to which subject. This information is inherently provided by the structure of the datasets. Many few-shot learning algorithms can be applied to this problem by considering the subjects to be classes with only a few samples. 

\subsection{Few-Shot Learning}

Few-Shot Learning is a subfield of machine learning focused on extracting meaningful features from limited samples per class, useful in data-scarce domains such as medical imaging. Problems are typically framed as $N$-way $K$-shot tasks, where $N$ denotes the number of classes and $K$ the number of samples per class. The main approaches include Transfer-Learning, Meta-Learning, and Metric-Learning. Transfer-Learning fine-tunes pre-trained models for new classes, Meta-Learning enables rapid adaptation from related tasks \cite{finn2017model}, and Metric-Learning optimizes distance metrics to cluster similar samples and separate dissimilar ones \cite{packhauser2022deep} \cite{BMVC2016_119}.\
This work adopts Metric-Learning for its simplicity and the absence of suitable $3D$ biomedical pre-trained models. The model minimizes Euclidean distances between embeddings of the same subject while maximizing distances between different subjects.

\section{Materials and Methods}

\subsection{Models}

In our experiments, we employ the $2D$ and $3D$ ResNet-50 \cite{DBLP:journals/corr/HeZRS15} \cite{hara2017learning} architectures, two popular and efficient neural network architectures well-suited to our problem setting. ResNet-50 is a convolutional neural network that utilizes residual blocks to mitigate the vanishing gradient problem. We use the final output of the model as an embedding for subject fingerprinting.

\subsection{Datasets}
\textbf{X-Ray:} For the $2D$ X-ray fingerprinting task, we utilized the NIH Chest X-ray Dataset (also known as \href{https://nihcc.app.box.com/v/ChestXray-NIHCC}{ChestXray-14}), published as an extension of the ChestXray-8 dataset \cite{wang2017chestxray} . This large-scale dataset contains $112'120$ frontal-view chest X-ray images from $30'805$ unique subjects collected between the years $1992$ and $2015$. The images show a large variety of image quality, subject position and posture, and up to fourteen common thoracic pathologies, making image augmentations, beyond normalization and resizing to a $512$ x $512$ resolution, unnecessary. We use the train-validation-test split from \cite{packhauser2022deep}. \\
\textbf{BraTS2021:} For the $3D$ MRI fingerprinting task, we utilized the Brain Tumor Segmentation (\href{http://braintumorsegmentation.org/}{BraTS}) 2021 dataset from the MICCAI BraTS Challenge \cite{DBLP:journals/corr/abs-2107-02314}, \cite{menze2014multimodal} and \cite{bakas2017advancing}. This multi-institutional dataset comprises $1'251$ subjects with brain tumors, each providing magnetic resonance imaging (MRI) scans with multiple modalities. Due to the limited natural variance in brain shape and position across modalities, we applied affine transformations to the images during training and resized the images to $78$ x $120$ x $120$. We randomly split the data on a patient-wise basis into a $70\%-10\%-20\%$ train-validation-test split.

\subsection{Training}


At each iteration step, we randomly select subjects from the train split and uniformly sample two images per subject to create a batch of anchor images and a batch of positive images. Following the work on Siamese Neural Networks \cite{bromley1993signature}, we encode the anchors and the positives using the same learnable model parameters. For each anchor-positive tuple, we search for an appropriate negative sample among the anchor images of distinct subjects within the batch. Finally, we use the triplet margin loss \cite{BMVC2016_119}, defined in \textit{Equation \ref{eq:triplet_margin_loss}}, to train the model by comparing the anchor embeddings, the positive and the negative samples. \\

\subsection{Triplet Margin Loss}

The triplet margin loss aims at minimizing the distance between the associated anchor $a$ and positive embedding $p$, pulling together the embeddings of the images of the same subject. However, relying only on this pulling force would lead to a model collapse. Therefore, we need an additional pushing force that maximizes the dissimilarity of embeddings coming from images of different subjects. To establish this, the triplet margin loss relies on a negative sample $n$ and maximizes the distance between the anchor and negative embeddings. The triplet margin loss encourages the distance between the anchor and a hard negative to be larger than the distance between the anchor and a positive by at least margin $\alpha$. A hard negative sample per anchor-positive tuple is selected from the batch of anchors. The triplet margin loss \cite{schroff2015facenet} is given as:

\begin{equation}
    \label{eq:triplet_margin_loss}
    \mathcal{L}_{trip}(a, p, n) = \max(d(a, p) - d(a, n) + \alpha, 0)
\end{equation}

Selecting random negatives from the batch is usually not enough for the network to learn meaningful embeddings. Negatives that are highly dissimilar to the anchor are easy-negatives and lead to a loss value of $0$. Therefore, we employ an in-batch mining strategy \cite{schroff2015facenet}, as described in \textit{Algorithm \ref{alg:online_triplet_mining_batch}}, to find hard negatives in the batch for each anchor-positive tuple. Due to this mining strategy, it is crucial that the batch size is large enough to include suitable hard negatives for the anchor-positive tuples.

\begin{algorithm}
\caption{Batched Hard Triplet Mining}
\label{alg:online_triplet_mining_batch}
\begin{algorithmic}[1]
\Require{Anchors ${a}$, Positives ${p}$, SubjectIDs ${id}$, Distance Function $d$}
\State $n \gets [\phantom{1}]$ \Comment{Negative indices}
\State bs = length(a) \Comment{Batch Size}
\For{$i = 1$ to $bs$}
  \State $d^+ = d(a_i, p_i)$ \Comment{Positive distance}
  \State $\mathcal{T}_{hard} \gets \emptyset$
  \For{$j = 1$ to $bs$}
    \If{$id_i \neq id_j$} \Comment{Validate subjects to be different}
        \State $d^- = d(a_i, a_j)$ \Comment{Negative Distance}
        \If{$d^- \leq d^+$}
            \State $\mathcal{T}_{hard}$.add($a_j$) \Comment{Hard Negative}
        \EndIf
    \EndIf
  \EndFor
  \If{$\mathcal{T}_{hard} \neq \emptyset$}
    \State n.append(random($\mathcal{T}_{hard}$)) \Comment{Random Hard Negative}
  \Else
    \State a.remove($a_i$) \Comment{No Suitable Negative Found}
    \State p.remove($p_i$)
  \EndIf
\EndFor
\State \Return $(a, p, n)$
\end{algorithmic}
\end{algorithm}

\subsection{Evaluation and Metrics}

We evaluate the models using an $N$-way $K$-shot framework, where $N$ represents the number of distinct subjects and $K$ denotes the number of support images available for each subject. For each subject, we assess the retrieval metric based on a single query image $q$, aiming to retrieve all its corresponding $K$ support images. This means that for each of the $N$ queries, we retrieve the subjects top $K$ most similar support images as a list $S_K =[s_1, \dots, s_K]$ ordered by the decreasing similarity scores. 

The Recall-@-$K$ ($Re_\varphi@K$) measures the percentage of relevant images that have been retrieved across all $N$ queries in the $N$-way $K$-shot problem $\varphi$. It is the ratio of query-relevant images retrieved, $TP_q$, to the total amount of relevant images $FN_q + TP_q = K$ averaged across all queries $N$.

\begin{equation}
    \label{eq:mean_recall_at_K_shot}
    Re_\varphi@K = \frac{1}{N} \sum_{q=1}^{N} \frac{TP_q}{TP_q + FN_q}= \frac{1}{N} \sum_{q=1}^{N} \frac{\sum_{s}^{S_K} rel(s, q)}{K}
\end{equation}

where,
\begin{equation*}
    rel(s, q) = \begin{cases} 
      1 & s = subject_q\\
      0 & \text{else} 
   \end{cases}
\end{equation*}

In our setting, the number of retrieved images ($TP_q + FP_q=K$) equals the number of relevant images ($TP_q+FN_q=K$), therefore the Recall at $K$ equals Precision at $K$.

Hit-@-R ($H_\varphi@R$) is an additional metric that checks if at least a single relevant image has been retrieved within the R most similar ranking images. It indicates how well relevant images rank compared to non-relevant images.

\begin{equation}
\label{eq:mean_hit_at_k}
    H_\varphi@R = \frac{1}{N} \sum_{q=1}^N (\bigvee_{s}^{S_R} rel(s, q))
\end{equation}

We repeat this experiment $100$ times with $N$ randomly sampled subjects and uniformly sample $K$ supports and one query. Finally, we average the metrics across runs.
\begin{equation}
    MRe@K = \frac{1}{100}\sum_{\varphi=1}^{100} Recall_\varphi@K
    \quad \text{and} \quad
    MH@R = \frac{1}{100}\sum_{\varphi=1}^{100} H_\varphi@R
\end{equation}

To assess embedding cluster quality, we calculate Mean Intra-Subject Distance (MIASD) and Mean Inter-Subject Distance (MIESD) as shown in \textit{Figure \ref{figure:histogram_and_figure} (A)}. MIASD measures how closely a subject's embeddings cluster around their mean embedding. MIESD measures the average distance between different subjects' mean embeddings. Together, these metrics reveal within-cluster tightness and between-cluster separation.

\section{Results}

\begin{table}[H]
    \centering
    \notsotiny
    \begin{minipage}[t]{0.55\textwidth}
        \centering
        \begin{tabular}{c||ccc||ccc||ccc}
            $N$-way - $K$-shot & \multicolumn{3}{c}{$MRe@K$} & \multicolumn{3}{c}{$MH@1$} & \multicolumn{3}{c}{$MH@5$} \\
            \hline
             & SSIM & Pack & Ours & SSIM & Pack & Ours & SSIM & Pack & Ours\\
            20-1  & 30.95 & 99.25 & 99.10 & 30.95 & 99.25 & 99.10 & 59.20 & 99.65 & 100.0\\
            20-5  & 27.24 & 99.57 & 98.24 & 49.55 & 99.85 & 99.70 & 73.05 & 99.90 & 99.95\\
            100-1 & 17.96 & 99.24 & 97.46 & 17.96 & 99.24 & 97.46 & 32.85 & 99.53 & 99.58\\
            100-5 & 15.91 & 99.20 & 95.33 & 33.05 & 99.81 & 99.17 & 51.12 & 99.86 & 99.72\\
            500-1 & 11.19 & 98.80 & 94.60 & 11.19 & 98.80 & 94.60 & 19.31 & 99.33 & 98.66\\
            500-5 &  9.82 & 98.55 & 90.06 & 23.03 & 99.66 & 98.05 & 35.25 & 99.53 & 99.44\\
            1000-1 & 9.55 & 98.51 & 92.30 & 9.55  & 98.51 & 92.30 & 16.08 & 99.18 & 97.95\\
        \end{tabular}
        \caption{ChestXray-14: SSIM, Pack and our performance on the test dataset across different $N$-way $K$-shot settings.}
        \label{table:xray_performance_assessment}
    \end{minipage}
    \hfill
    \begin{minipage}[t]{0.385\textwidth}
        \centering
        \begin{tabular}{c||c||c||c}
            $N$-way - $K$-shot & $MRe@K$ & $MH@1$ & $MH@3$ \\
            \hline
            20-1 & 99.20 &  99.20 & 99.65\\
            20-3 & 99.25 & 99.72 & 99.78\\
            100-1 & 98.86 & 98.86 & 99.25\\
            100-3 & 98.79 & 99.44 & 99.53\\
        \end{tabular}
        \caption{BraTS-2021: our ResNet-50 performance on the test dataset across different $N$-way $K$-shot settings.}
        \label{table:brats_performance_assessment}
    \end{minipage}
\end{table}

On the ChestXray-14 dataset, we compared the retrieval performances of our method to two other approaches (Table \ref{table:xray_performance_assessment}): \textbf{SSIM} \cite{nilsson2020understandingssim}, a well-established similarity measure, is used to compare pairs of images. The images are downsampled to $256x256$ resolution due to the slow nature of comparing images on a pixel basis. Similar to our method, \cite{Packh_user_2022} \textbf{(PACK)} compares the image embeddings of a ResNet-50 and is trained using a margin-based contrastive loss, however, they additionally make use of cross-batch memory \cite{DBLP:journals/corr/abs-1912-06798}. In Table \ref{table:xray_performance_assessment}, we can observe that for all models the $MRe@K$ performance drops as the amount of subjects increase, as it is more likely to include images from other subjects that are similar to the query images. On the other hand, the 5-shot settings show better $MH@1$ and $MH@5$ scores as more images from the same subject in the set of all possible images increase the possibility to include an image that is more similar to the query. Similarly, on the BraTS-2021 dataset, our method achieves a high performance on all metrics and the scores slightly decay with increasingly harder test settings. On ChestXray-14, Pack consistent outperforms our method, which we conclude to originate from two main differences: Pack is trained on higher resolution images ($1024x1024$) also used during retrieval. Using cross-batch memory, Pack can keep more negative samples in memory increasing the amount of good anchor-negative pairs during training.

\begin{figure}[H]
    \centering
    \includegraphics[scale=0.35]{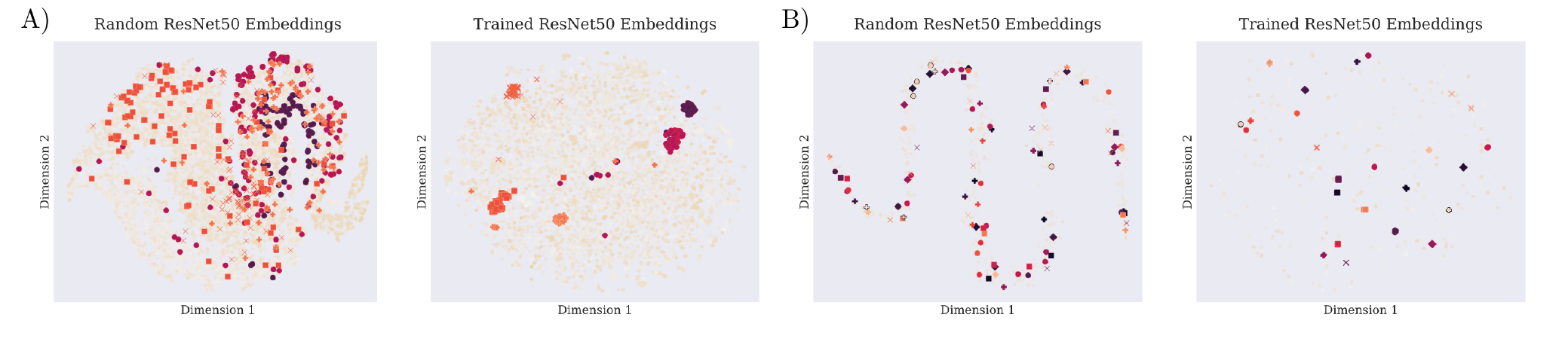}
    \caption{T-SNE plots of 128-dimensional image embeddings from the ChestXray-14 (A) and the BraTS-2021 (B) test sets, highlighting a selection of subjects. (A \& B left) Untrained ResNet-50 encodings, showing no clustering of subjects. (A \& B right) ResNet-50 trained with triplet margin loss, showing clustered and well-separated subjects in latent space.}
    \label{fig:tsne_plots_3D_and_2D}
\end{figure}

\begin{figure}[H]
    \centering
    \includegraphics[scale=0.24]{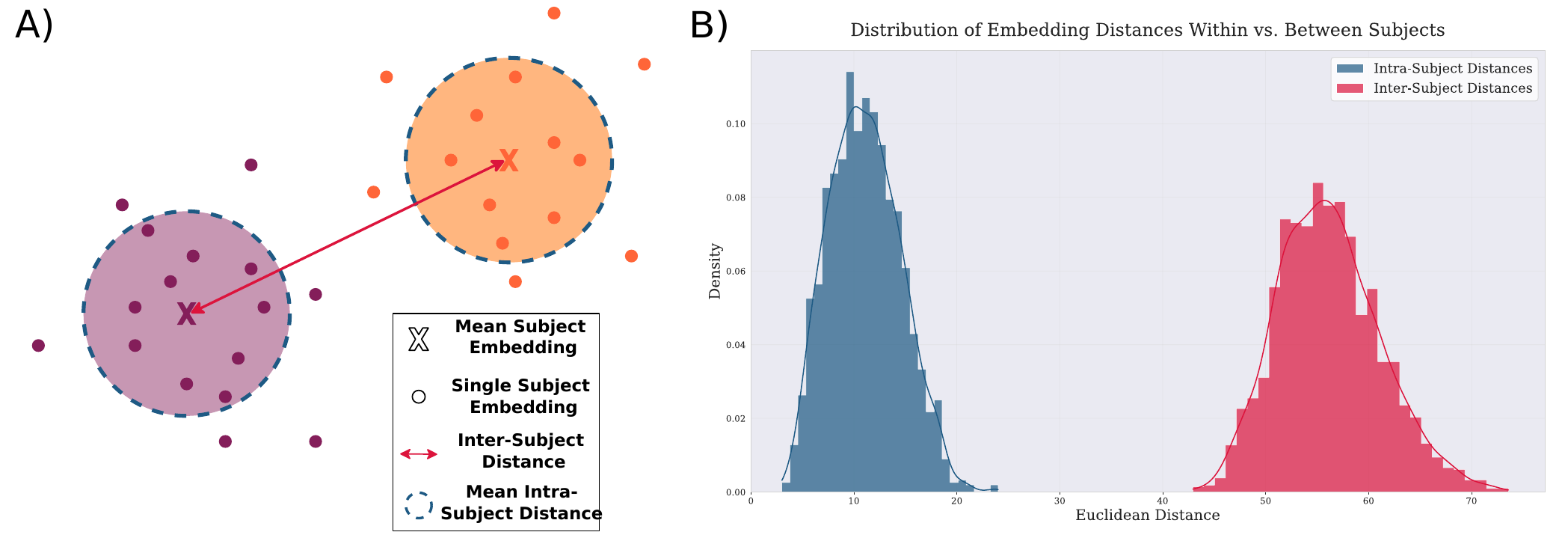}
    \caption{(A) Abstraction of the MIASD and MIESD. B) Binned Intra-Subject (Blue) and Inter-Subject Distances (Red) for all subjects in the ChestXray-14 dataset.}
    \label{figure:histogram_and_figure}
\end{figure}

In \textit{Figure \ref{fig:tsne_plots_3D_and_2D}}, we qualitatively show the models' ability to generate embeddings that form distinct clusters for images of the same subject within the latent space and effectively separate different subjects on the BraTS-2021 and ChestXray-14 datasets. In \textit{Figure \ref{figure:histogram_and_figure} (B)}, we quantitatively asses the clustering performance of the models using the MIESD and MIASD metrics. On ChestXray-14, the model achieved a MIASD value of $11.08 \pm 4.96$, and a MIESD value of $56.17 \pm 3.46$, On BraTS-2021, the model generates embeddings with a MIASD value of $3.01 \pm 1.97$ and a MIESD value of
$52.03 \pm 8.17$. This shows a clear separation of distinct subjects while clustering the images of the same subject in latent space.


\section{Discussion}

In conclusion, our simple training strategy demonstrates to be highly capable of differentiating and fingerprinting both $2D$ and $3D$ medical subject images. However, these results could be further improved by adding strategies circumventing memory limitations, such as cross-batch memory for contrastive learning. To the best of our knowledge, this is the first work using $3D$ volumetric biomedical data as input to a model for subject fingerprinting, and performing cross-modal MRI subject retrieval. We show that images form the same subject cluster together, while clusters of different subjects are well separated. This identification mechanism makes our approach particularly valuable for detecting near-duplicate images in both $2D$ and $3D$ medical image datasets, thereby providing a novel approach to mitigate data leakage when aggregating datasets in clinical research and machine learning applications.

\section{Code and Data}
The code can be found in the following repository {\small\textcolor{blue}{\url{https://github.com/BenBausch/Few-Shot-Fingerprinting}}} and the data can be downloaded unter the following links: NIH Chest X-rays {\small\textcolor{blue}{\url{https://www.kaggle.com/datasets/nih-chest-xrays/data}}} and BraTS2021 {\small\textcolor{blue}{\url{https://www.kaggle.com/datasets/dschettler8845/brats-2021-task1/data}}}.

\section{Acknowledgments}
The results presented in this article were obtained using the HPC facilities of the University of Luxembourg \cite{varrette2022management} (https://hpc.uni.lu).

\newpage
\printbibliography

@article{DBLP:journals/corr/HeZRS15,
  author       = {Kaiming He and
                  Xiangyu Zhang and
                  Shaoqing Ren and
                  Jian Sun},
  title        = {Deep Residual Learning for Image Recognition},
  journal      = {CoRR},
  volume       = {abs/1512.03385},
  year         = {2015},
  url          = {http://arxiv.org/abs/1512.03385},
  eprinttype    = {arXiv},
  eprint       = {1512.03385},
  bibsource    = {dblp computer science bibliography, https://dblp.org}
}

@inproceedings{BMVC2016_119,
        	title={Learning local feature descriptors with triplets and shallow convolutional neural networks},
        	author={Vassileios Balntas, Edgar Riba, Daniel Ponsa and Krystian  Mikolajczyk},
        	year={2016},
        	month={September},
        	pages={119.1-119.11},
        	articleno={119},
        	numpages={11},
        	booktitle={Proceedings of the British Machine Vision Conference (BMVC)},
        	publisher={BMVA Press},
        	editor={Richard C. Wilson, Edwin R. Hancock and William A. P. Smith},
        	doi={10.5244/C.30.119},
        	isbn={1-901725-59-6},
        	url={https://dx.doi.org/10.5244/C.30.119}
}

@InProceedings{wang2017chestxray,
author = {Wang, Xiaosong and Peng, Yifan and Lu, Le and Lu, Zhiyong and Bagheri, Mohammadhadi
and Summers, Ronald},
title = {ChestX-ray8: Hospital-scale Chest X-ray Database and Benchmarks on Weakly-Supervised
Classification and Localization of Common Thorax Diseases},
booktitle = {2017 IEEE Conference on Computer Vision and Pattern Recognition (CVPR)},
pages = {3462--3471},
year = {2017}
}

@article{DBLP:journals/corr/abs-2107-02314,
  author       = {Ujjwal Baid and
                  Satyam Ghodasara and
                  Michel Bilello and
                  Suyash Mohan and
                  Evan Calabrese and
                  Errol Colak and
                  Keyvan Farahani and
                  Jayashree Kalpathy{-}Cramer and
                  Felipe C. Kitamura and
                  Sarthak Pati and
                  Luciano M. Prevedello and
                  Jeffrey D. Rudie and
                  Chiharu Sako and
                  Russell T. Shinohara and
                  Timothy Bergquist and
                  Rong Chai and
                  James A. Eddy and
                  Julia Elliott and
                  Walter Reade and
                  Thomas Schaffter and
                  Thomas Yu and
                  Jiaxin Zheng and
                  BraTS Annotators and
                  Christos Davatzikos and
                  John Mongan and
                  Christopher Hess and
                  Soonmee Cha and
                  Javier E. Villanueva{-}Meyer and
                  John B. Freymann and
                  Justin S. Kirby and
                  Benedikt Wiestler and
                  Priscila Crivellaro and
                  Rivka R. Colen and
                  Aikaterini Kotrotsou and
                  Daniel S. Marcus and
                  Mikhail Milchenko and
                  Arash Nazeri and
                  Hassan M. Fathallah{-}Shaykh and
                  Roland Wiest and
                  Andr{\'{a}}s Jakab and
                  Marc{-}Andr{\'{e}} Weber and
                  Abhishek Mahajan and
                  Bjoern H. Menze and
                  Adam E. Flanders and
                  Spyridon Bakas},
  title        = {The {RSNA-ASNR-MICCAI} BraTS 2021 Benchmark on Brain Tumor Segmentation
                  and Radiogenomic Classification},
  journal      = {CoRR},
  volume       = {abs/2107.02314},
  year         = {2021},
  url          = {https://arxiv.org/abs/2107.02314},
  eprinttype    = {arXiv},
  eprint       = {2107.02314},
  bibsource    = {dblp computer science bibliography, https://dblp.org}
}

@article{menze2014multimodal,
  title={The multimodal brain tumor image segmentation benchmark (BRATS)},
  author={Menze, Bjoern H and Jakab, Andras and Bauer, Stefan and Kalpathy-Cramer, Jayashree and Farahani, Keyvan and Kirby, Justin and Burren, Yuliya and Porz, Nicole and Slotboom, Johannes and Wiest, Roland and others},
  journal={IEEE transactions on medical imaging},
  volume={34},
  number={10},
  pages={1993--2024},
  year={2014},
  publisher={IEEE}
}

@article{bakas2017advancing,
  title={Advancing the cancer genome atlas glioma MRI collections with expert segmentation labels and radiomic features},
  author={Bakas, Spyridon and Akbari, Hamed and Sotiras, Aristeidis and Bilello, Michel and Rozycki, Martin and Kirby, Justin S and Freymann, John B and Farahani, Keyvan and Davatzikos, Christos},
  journal={Scientific data},
  volume={4},
  number={1},
  pages={1--13},
  year={2017},
  publisher={Nature Publishing Group}
}

@inproceedings{finn2017model,
  title={Model-agnostic meta-learning for fast adaptation of deep networks},
  author={Finn, Chelsea and Abbeel, Pieter and Levine, Sergey},
  booktitle={International conference on machine learning},
  pages={1126--1135},
  year={2017},
  organization={PMLR}
}

@article{bromley1993signature,
  title={Signature verification using a" siamese" time delay neural network},
  author={Bromley, Jane and Guyon, Isabelle and LeCun, Yann and S{\"a}ckinger, Eduard and Shah, Roopak},
  journal={Advances in neural information processing systems},
  volume={6},
  year={1993}
}

@inproceedings{hara2017learning,
  title={Learning spatio-temporal features with 3d residual networks for action recognition},
  author={Hara, Kensho and Kataoka, Hirokatsu and Satoh, Yutaka},
  booktitle={Proceedings of the IEEE international conference on computer vision workshops},
  pages={3154--3160},
  year={2017}
}

@article{packhauser2022deep,
  title={Deep learning-based patient re-identification is able to exploit the biometric nature of medical chest X-ray data},
  author={Packh{\"a}user, Kai and G{\"u}ndel, Sebastian and M{\"u}nster, Nicolas and Syben, Christopher and Christlein, Vincent and Maier, Andreas},
  journal={Scientific Reports},
  volume={12},
  number={1},
  pages={14851},
  year={2022},
  publisher={Nature Publishing Group UK London}
}

@inproceedings{schroff2015facenet,
  title={Facenet: A unified embedding for face recognition and clustering},
  author={Schroff, Florian and Kalenichenko, Dmitry and Philbin, James},
  booktitle={Proceedings of the IEEE conference on computer vision and pattern recognition},
  pages={815--823},
  year={2015}
}

@article{macpherson2023patient,
  title={Patient reidentification from chest radiographs: an interpretable deep metric learning approach and its applications},
  author={Macpherson, Matthew S and Hutchinson, Charles E and Horst, Carolyn and Goh, Vicky and Montana, Giovanni},
  journal={Radiology: Artificial Intelligence},
  volume={5},
  number={6},
  pages={e230019},
  year={2023},
  publisher={Radiological Society of North America}
}

@INPROCEEDINGS{Truong2024Benchmarking,
  author={Truong, Tuan and Jush, Farnaz Khun and Lenga, Matthias},
  booktitle={2024 IEEE International Symposium on Biomedical Imaging (ISBI)}, 
  title={Benchmarking Pretrained Vision Embeddings for Near- and Duplicate Detection in Medical Images}, 
  year={2024},
  volume={},
  number={},
  pages={1-5},
  doi={10.1109/ISBI56570.2024.10635550}}

@article{Packh_user_2022,
   title={Deep learning-based patient re-identification is able to exploit the biometric nature of medical chest X-ray data},
   volume={12},
   ISSN={2045-2322},
   url={http://dx.doi.org/10.1038/s41598-022-19045-3},
   DOI={10.1038/s41598-022-19045-3},
   number={1},
   journal={Scientific Reports},
   publisher={Springer Science and Business Media LLC},
   author={Packhäuser, Kai and Gündel, Sebastian and Münster, Nicolas and Syben, Christopher and Christlein, Vincent and Maier, Andreas},
   year={2022},
   month=sep }

@article{DBLP:journals/corr/abs-1912-06798,
  author       = {Xun Wang and
                  Haozhi Zhang and
                  Weilin Huang and
                  Matthew R. Scott},
  title        = {Cross-Batch Memory for Embedding Learning},
  journal      = {CoRR},
  volume       = {abs/1912.06798},
  year         = {2019},
  url          = {http://arxiv.org/abs/1912.06798},
  eprinttype    = {arXiv},
  eprint       = {1912.06798},
  bibsource    = {dblp computer science bibliography, https://dblp.org}
}

@misc{nilsson2020understandingssim,
      title={Understanding SSIM}, 
      author={Jim Nilsson and Tomas Akenine-Möller},
      year={2020},
      eprint={2006.13846},
      archivePrefix={arXiv},
      primaryClass={eess.IV},
      url={https://arxiv.org/abs/2006.13846}, 
}

@inproceedings{varrette2022management,
  title={Management of an academic HPC \& research computing facility: The ULHPC experience 2.0},
  author={Varrette, Sebastien and Cartiaux, Hyacinthe and Peter, Sarah and Kieffer, Emmanuel and Valette, Teddy and Olloh, Abatcha},
  booktitle={Proceedings of the 2022 6th High Performance Computing and Cluster Technologies Conference},
  pages={14--24},
  year={2022}
}
\end{document}